\documentclass[letterpaper, 10 pt, conference]{ieeeconf}  

\IEEEoverridecommandlockouts                              

\usepackage{hyperref}
\usepackage{graphicx}
\usepackage{amsmath}
\usepackage{cite}
\usepackage{algorithm}
\usepackage{algpseudocode}
\usepackage{multirow}
\usepackage{gensymb}
\usepackage{booktabs}
\usepackage{siunitx}

\title{\LARGE \bf{LLM-Craft: Robotic Crafting of Elasto-Plastic Objects with Large Language Models}}

\author{Alison Bartsch$^{1}$, and Amir Barati Farimani$^{1}$
\thanks{$^{1}$With the Department of Mechanical Engineering,
        Carnegie Mellon University \tt\small \{abartsch, afariman\} @andrew.cmu.edu}}

\begin{document}

\maketitle
\thispagestyle{empty}
\pagestyle{empty}






\begin{abstract}

When humans create sculptures, we are able to reason about how geometrically we need to alter the clay state to reach our target goal. We are not computing point-wise similarity metrics, or reasoning about low-level positioning of our tools, but instead determining the higher-level changes that need to be made. In this work, we propose LLM-Craft, a novel pipeline that leverages large language models (LLMs) to iteratively reason about and generate deformation-based crafting action sequences. We simplify and couple the state and action representations to further encourage shape-based reasoning. To the best of our knowledge, LLM-Craft is the first system successfully leveraging LLMs for complex deformable object interactions. Through our experiments, we demonstrate that with the LLM-Craft framework, LLMs are able to successfully create a set of simple letter shapes. We explore a variety of rollout strategies, and compare performances of LLM-Craft variants with and without an explicit goal shape images. For videos and prompting details, please visit our project website: \href{https://sites.google.com/andrew.cmu.edu/llmcraft/home}{https://sites.google.com/andrew.cmu.edu/llmcraft/home}.

\end{abstract}

\section{Introduction}

    With the goal of building robust and generalizable autonomous robotic systems, we need to design frameworks that can reason about complex interactions between the robot and the environment. Deformable object manipulation is a great test task for reasoning about these complex interactions, as it requires understanding how the state of the object itself will change during contact. In this work, we will investigate the task of deformation-based crafting, in which the robot molds the clay into a variety of simple shapes with a parallel gripper. This task allows us to investigate the challenges of both predicting the behavior of the clay as well as long-horizon planning and reasoning. Previous work focused on elasto-plastic object shaping typically work at the very low-level, predicting clay dynamics \cite{shi2022robocraft, bartsch2023sculptbot}, leveraging human-created sub-goals to speed up action planning \cite{shi2023robocook} or direct action imitation \cite{bartsch2024sculptdiff}. However, these approaches do not leverage existing world knowledge to reason about how clay behaves or have the ability to reason at a geometric or semantic level. Furthermore, existing methods for the clay shaping task rely heavily on hardware-specific real-world datasets which can be very time consuming to collect and are often not transferable across hardware configurations \cite{shi2022robocraft, shi2023robocook, bartsch2023sculptbot, bartsch2024sculptdiff, bauer2024doughnet}.


    In this work, we aim to explore to what extent we are able to directly leverage Large Language Models (LLMs) without any task-specific data or finetuning for this highly complex manipulation task. Particularly, do LLMs contain useful world knowledge to generate long-horizon action sequences to create simple letter shapes with deformation-based actions? LLMs have been demonstrated to contain useful world knowledge for a wide range of reasoning and generation tasks \cite{mirchandani2023large, wu2023smartplay, todd2023level, yu2023language, jadhav2024large, wang2024can}. However, past work using LLMs to generate plans for robotic tasks typically focus on simple pick and place actions that require less sophisticated prediction of future states \cite{singh2023progprompt, sharan2024plan, wu2024mldt, bhat2024grounding, gupta2024action}. We present LLM-Craft, a preliminary study exploring how LLMs reason about the complex task of elasto-plastic object crafting. We hypothesize that LLMs have substantial and useful world knowledge to generate action sequences for the clay crafting task, such as understanding how clay will behave when pressed or squeezed and understanding simple semantic shapes and goals. To better facilitate higher-level geometric reasoning for crafting action selection, we simplify our state and action representation substantially. Through extensive real-world experiments, we demonstrate the success of leveraging LLMs directly for the clay crafting task. We find that the ability of LLMs to reason at the geometric and semantic level is particularly powerful and presents an argument for future crafting pipelines incorporating LLMs.

\begin{figure*}
    \centering
    \includegraphics[width=0.99\linewidth]{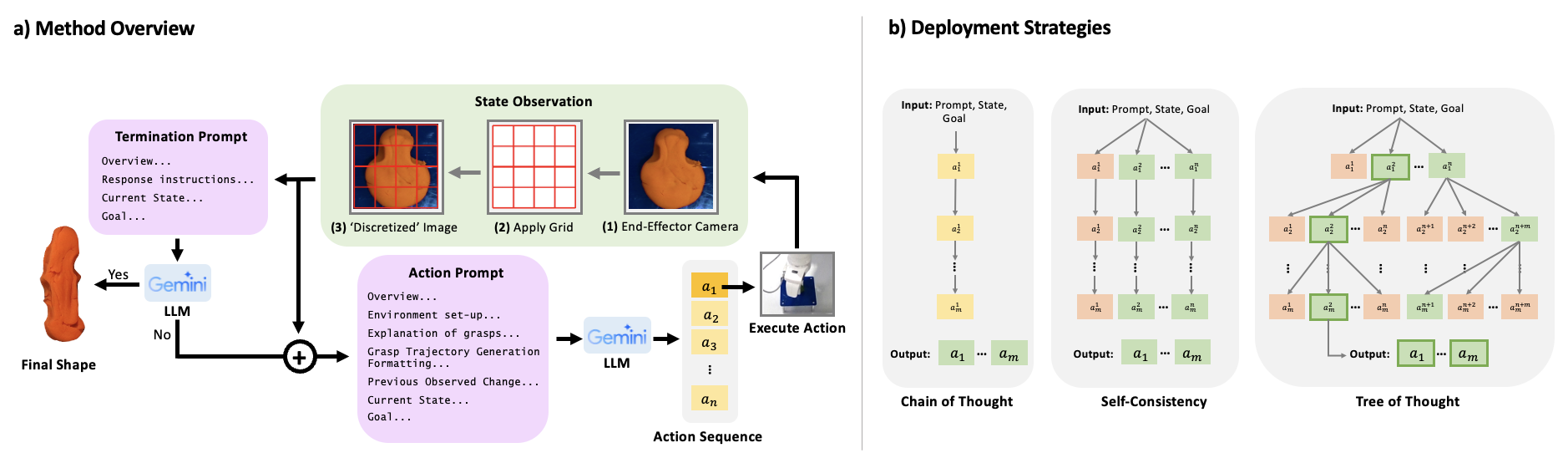}
    \caption{\textbf{Method Pipeline.} \textbf{a)} The LLM-Craft system takes a top-down image of the clay with a wrist-mounted camera as the state observation. A grid is applied to the image to represent the discrete regions of the clay. The LLM is prompted with the gridded state/goal images and an action prompt. The LLM selects a sequence of grasps to apply to the clay, and the robot executes the first one. A new state observation is collected and passed back to the LLM along with the goal image and the termination propmt to determine if the goal has been reached. If the goal has not been reached, the LLM is iteratively re-queried until the goal has been achieved. \textbf{b)} A visualization of different method deployment strategies. We present chain of thought \cite{wei2022chain} as our iterative base method, and self-consistency (SC) \cite{wang2022self} and tree of thought (ToT) \cite{yao2023tree} are rollout strategies that require multiple queries of the LLM per action trajectory generation.}
    \label{fig:methods}
\end{figure*}

\section{Related works}

    \textbf{Elasto-Plastic Object Manipulation: } There have been a variety of successful past works within the realm of elasto-plastic object manipulation and crafting. There are numerous simulations modeling the deformation behavior of these objects, but there remains a large sim-to-real gap \cite{heiden2021disect, hu2018moving, huang2021plasticinelab, gu2023maniskill2}. These simulations enabled trajectory optimization-based approaches that were able to bridge the sim-to-real gap for simple tasks \cite{qi2022learning, yamada2024d}. Many real-world clay crafting frameworks learn the low-level dynamics of the clay to be used to generate action trajectories \cite{shi2022robocraft, shi2023robocook, bartsch2023sculptbot, bauer2024doughnet, li2024deformnet}. Alternatively, a range of imitation-based frameworks have been successful, where demonstrations are abstracted into skills \cite{lin2022planning, li2023dexdeform}, or more direct trajectory imitation \cite{bartsch2024sculptdiff}. All of these successful real-world approaches require hardware and task-specific datasets, with many requiring human demonstrations which are particularly time consuming to collect. Additionally, these methods are very brittle to particular design choices and are not always transferable across lab setups. In this work, we investigate if we can leverage the general world knowledge of LLMs to shape elasto-plastic objects with a parallel gripper. From the lens of the deformable manipulation community, if LLMs do indeed have useful world knowledge to make high-level predictions of the objects' behavior, this could be a useful component to future pipelines that could improve generalizability.

    \textbf{LLMs for Robotic Tasks: } There is a significant body of previous work leveraging LLMs for a variety of robotic tasks. LLMs have been particularly effective as robotic task planning agents \cite{wu2024mldt, wang2023describe, bhat2024grounding, rana2023sayplan, dorbala2023can, sharan2024plan, singh2023progprompt, ding2023task, gupta2024action,nasiriany2024pivot, cheng2024empowering} with many recent works finding success incorporating a variety of real-world grounding techniques including scene graphs \cite{rana2023sayplan}, diffusion models \cite{sharan2024plan}, counterfactual perturbation \cite{wang2024grounding}, developing a multimodal language model for embodied tasks \cite{driess2023palm} and 3D value maps \cite{huang2023voxposer}. These LLM-based frameworks span a wide range of reasoning levels, from low-level autoregressive code policy generation \cite{liang2023code} to reasoning about task plans purely with natural language \cite{mikami2024natural}. While the widespread success of LLM robotic methods has been impressive, most research has focused on the relatively simple tasks of navigation \cite{dorbala2023can, shah2023lm, rana2023sayplan, yang2023llm}, or sequences of pick and place actions \cite{singh2023progprompt, sharan2024plan, wu2024mldt, bhat2024grounding, gupta2024action}. These tasks do require long-horizon reasoning as well as recognition and interaction with a variety of real-world objects, but the general action skillset is relatively simple and does not require much intuition and understanding of physical robot-object interactions. In Tidybot \cite{wu2023tidybot}, researchers present an LLM planning framework that involves interacting with deformable cloth, but the specific sorting tasks explored do not require reasoning about the complex deformable dynamics and the robot-object interactions remain only pick and place. Similarly, in \cite{xu2023creative} researchers explore how LLMs can creatively reason about tool use to achieve a set of tasks, but the tool-object interactions remain very simple. In fact, in \cite{kwon2023language} researchers concluded that LLMs often simplify or incorrectly predict more complicated interactions between the robot and objects. In this work, we hypothesize that LLMs contain useful world knowledge to generate coherent sculpting sequences, such as needing to conduct sequential perpendicular grasps to create an \textit{X} shape. In this work, we explore how LLMs can plan action sequences involving complex robot-object interactions by predicting an action sequence as well as the predicted effect of the grasp action on the objects' state itself.

\section{Methodology}
\label{sec:method}

The key components of LLM-Craft are the choice of state and action representations as well as the iterative prompting scheme with the LLM. An overview of the pipeline is shown in Figure \ref{fig:methods}. In this work, we are using Gemini 1.0 Pro Vision \cite{team2023gemini} and Gemini 2.0 Flash \cite{Pichai_Hassabis_Kavukcuoglu_2024} as our multimodal large language models that can take both images and text as input. Due to the fast-moving nature of state-of-the-art LLMs, model versions were depreciated through our experiment revisions. We have done our best to allow for comparison between model versions and experiment ablations. We have not investigated alternative LLMs, but the LLM-Craft pipeline can be used with any LLM that can take both images and text as input, such as GPT-4 \cite{achiam2023gpt}.

\subsection{State and Action Representation}

In this work we are proposing an LLM-based framework for the task of top-down shape crafting of elasto-plastic objects with a parallel gripper. In existing robotic clay crafting literature, the common choice of state representation is a point cloud, and the action representation is the continuous end-effector pose \cite{shi2022robocraft, bartsch2023sculptbot, shi2023robocook, bartsch2024sculptdiff}. While these state and action choices are very effective in existing pipelines, they can disconnect the reasoning about actions from the state representation. When reasoning about how to shape clay, humans determine at a high level where and how the clay needs to be deformed. In this work, we propose a simple gridded top-down image of the clay as the state and action space to allow the LLM to reason about the larger geometric regions of the clay without needing to directly control global end-effector poses. The gridded state space is visualized in Figure \ref{fig:methods}. We equip the robot with an end-effector mounted camera to capture top-down images of the clay. We then superimpose a simple 4x4 grid onto the image to break the clay into different geometrical regions. Each cell in the grid is 2cm by 2cm in the robot frame, as the 3D printed fingertips on the robot are 2cm in width. A visualization of the hardware setup is shown in Figure \ref{fig:hardware}. We represent the columns of the grid from left to right as 'A', 'B', 'C', and 'D' and the rows from top to bottom as '1', '2', '3', and '4', thus cells individually are referenced as 'A1, 'B1', etcetera. The action space is defined as selecting two cells to push a distance towards each other. This choice of action representation significantly simplifies action selection as there is a discrete set of actions for the LLM to choose from, which is a much simpler task than selecting continuous action parameters. Additionally, this action representation is more closely linked to predicting the clay deformation behavior, than directly selecting the global end-effector pose. Through experiments we find that while this representation drastically constrains the action space, both the LLM-Craft and a human baseline are able to successfully create a variety of shape goals. At execution time, we assume the clay position remains fixed on the elevated stage, and thus know the global position of each of the grid points to transform the discrete cells selected by the LLM into an end-effector pose for squeezing.

\subsection{Prompting}

In this work, we explore how we can leverage the world knowledge and shape-level reasoning capabilities of LLMs without requiring any data collection, further training or fine-tuning on our specific task. The full process diagram of the iterative prompting scheme is shown in Figure \ref{fig:methods}, in which we use an action prompt and a termination prompt. The action prompt provides the LLM with an explanation of the task, state/action space, and the robot embodiment, and requires the LLM to output a crafting action sequence of varying length. We take inspiration from past successful approaches using LLMs for robotic tasks for some of the prompting details \cite{kwon2023language}. The components of the action prompt are as follows: \textbf{(1)} an overview of the robot embodiment and the high-level details of the crafting task, \textbf{(2)} the environmental setup including the clay and the grid coordinate system, \textbf{(3)} the grasp action, including that the end-effector pose cannot change during the grasp, \textbf{(4)} how to generate a crafting trajectory, including the requirements of (a) describing the current and goal shapes, (b) reasoning about the grasp strategy, (c) describe the predicted effect of the chosen grasp to the clay state, \textbf{(5)} the goal gridded image, \textbf{(6)} the current state gridded image, and \textbf{(7)} the user command to create the goal from the current state. The role of the termination prompt is to query the LLM to determine if the goal shape has been reached. The termination prompt consists of \textbf{(1)} an overview of the shape comparison task, \textbf{(2)} how to generate a decision, with the requirements to (a) describe the similarities and differences between the state and goal, and (b) make a decision if crafting should continue, \textbf{(3)} the current gridded state image, \textbf{(4)} the goal gridded image, and \textbf{(5)} the user command to determine if crafting should stop. For full details of both prompts, please see the Supplemental Materials on the project website. In section \ref{sec:ablation} we will discuss prompt ablations to evaluate the impactfullness of each component of the prompts on the final system performance.

\subsection{Rollout Strategies}

In our experiments, we explore standard iterative querying of the LLM with Chain of Thought prompting \cite{wei2022chain}, as well as more sophisticated strategies to mitigate error, such as Self-Consistency (SC) \cite{wang2022self} and Tree of Thought (ToT) \cite{yao2023tree}. For our Self-Consistency implementation, we leverage sampling-based generation in which we query the LLM 10 times to generate candidate sculpting trajectories, and execute the most frequently generated trajectory. Due to the symmetrical nature of our squeezing actions, we consider 'squeeze A1 and B2' and 'squeeze B2 and A1' to be the same actions. For our breadth-first-search Tree of Thought implementation, we leverage proposal-based generation in which we prompt the LLM to propose 4 possible actions given the current state and action history. This better ensures that we generate a wide diversity of candidate sculpting trajectories without generating any duplicates. Additionally, we employ voting-based value prediction of each candidate trajectory in the tree, as it is very difficult to quantify the value of candidate sculpting trajectories directly. 

\begin{figure}
        \centering
        \includegraphics[width=0.85\linewidth]{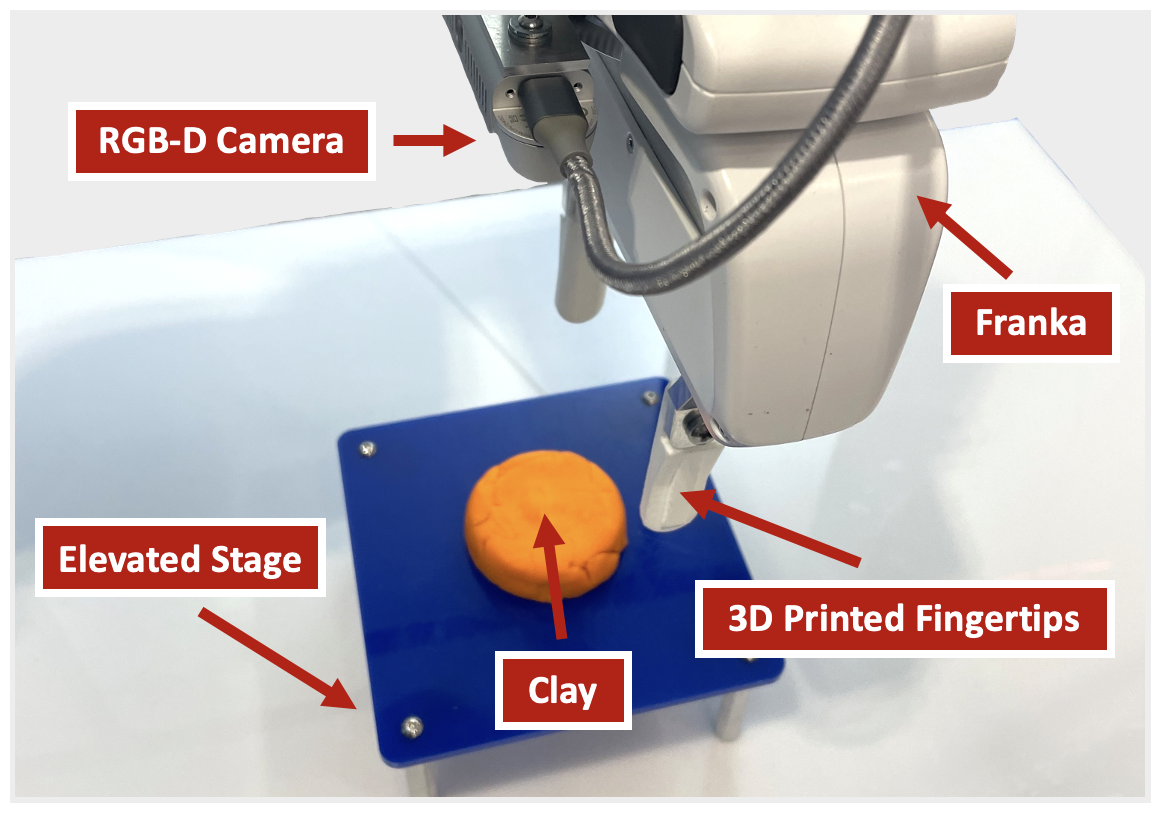}
        \caption{\textbf{Experimental Setup.} The experimental setup consists of a wrist-mounted Intel RealSense D415 camera, 3D printed fingertips for the parallel gripper, and an elevated stage for the clay.}
        \label{fig:hardware}
\end{figure}

\section{Experiments}
\label{sec:result}

    To fully evaluate LLM-Craft we conduct a variety of experiments, particularly with single-step shape goals and longer horizon letter shape goals evaluating the system performance as well as the impacts of the choice of state representation, grid size, image versus text-based goals, and the rollout strategy. Additionally, we conduct an ablation study to evaluate the importance of each component of the prompt on the final performance, as well as the impact of varying grid sizes. For each experiment, we conduct five real-world runs per goal and report the mean and standard deviation of all quantitative metrics. The hardware setup is shown in Figure \ref{fig:hardware}. With a wrist-mounted RGB-D camera, we record the top-down image of the clay before and after each grasp. Additionally after the crafting sequence is complete, we record a top-down point cloud of the clay for our quantitative similarity metric evaluations compared to a point cloud of the goal shape. 

    \subsection{Quantitative Evaluation Metrics}

    Chamfer Distance (CD) and Earth Mover's Distance (EMD) between the final point cloud and the goal shape are the most common quantitative metrics used to evaluate the clay shaping task in literature \cite{shi2023robocook, shi2022robocraft, bartsch2023sculptbot, bartsch2024sculptdiff}. However, for the more semantic task of shaping a specific letter, CD and EMD do not fully capture the nuanced qualities humans associate with a 'good' clay shape. Therefore, in addition to reporting CD/EMD, we propose additional metrics to represent the curvature smoothness of the shapes created. We represent the "smoothness" of the created shapes with the equation of curvature of parametric curves (equation \ref{eq:curve}), and with the perimeter-to-area ratio (PAR) on the 2D outline of each shape extracted with Canny edge detection.

    \begin{equation}
    \label{eq:curve}
        curvature = \frac{||x''y' - y''x'||}{(x'^2 + y'^2)^\frac{3}{2}}
    \end{equation}

    Beyond these quantitative metrics, we conducted a human evaluation study following the procedure of RoboCook \cite{shi2023robocook} to further evaluate the performance of LLM-Craft. To evaluate all 11 methods (in \ref{tab:general_performance}) over the 5 different shapes (\textit{C, I, L, T, X}), we created 11 randomized surveys each with a single instance of each letter and a random method. This ensured that each survey taker only saw a single instance of each letter, as to not bias their responses. We asked each respondent to first classify the letter in the provided image, and then rate the shape quality on a scale from 1-10.

    \begin{table}
    \caption{\textbf{State Representation Quantitative Results}. We present results across 5 single-step goals, with 5 runs per goal.}\label{tab:single_step_performance} 
    \centering
    \begin{tabular}{@{\extracolsep{\fill}}lll}
            \hline
            \textbf{State} & \textbf{CD $\downarrow$} & \textbf{EMD $\downarrow$}   \\ 
            \hline
            \hline
             Image & \textbf{0.0036 $\pm$ 0.0010} & \textbf{0.0061 $\pm$ 0.0022}   \\
             Array & 0.0037 $\pm$ 0.0012 & 0.0061 $\pm$ 0.0025  \\
            \hline
        \end{tabular}
    \end{table}

    \begin{figure}
    \centering
    \includegraphics[width=0.6\linewidth]{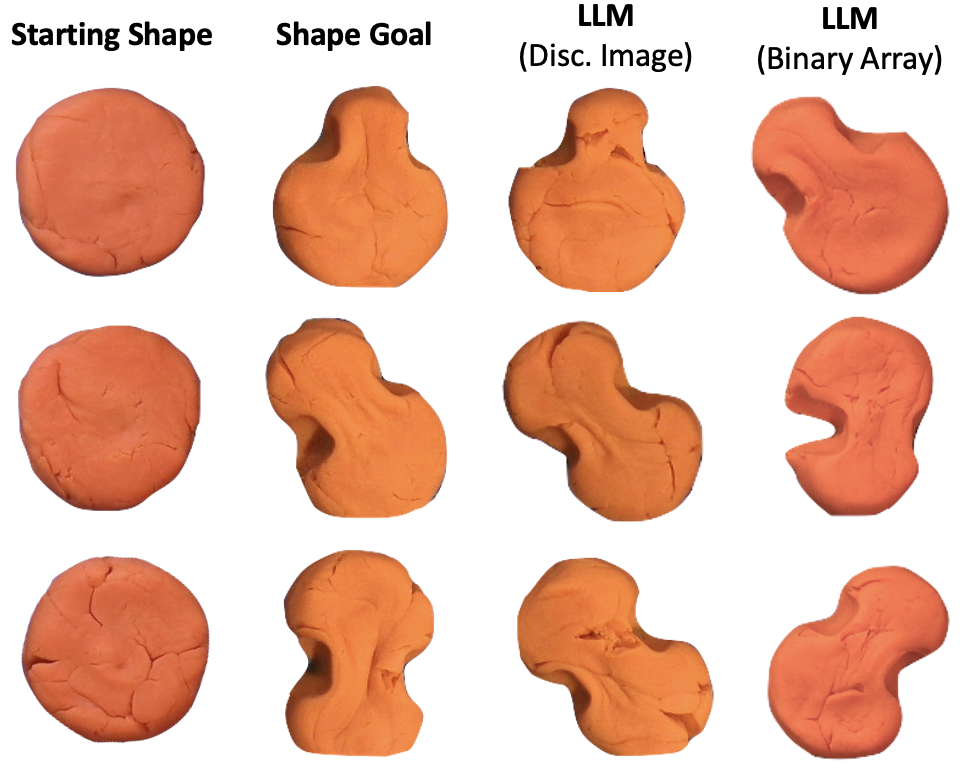}
    \caption{\textbf{Qualitative Single-Step Results.} We find for the single-step task without a semantically meaningful goal, the LLM can struggle with rotation in both cases of state representation.}
    \label{fig:single_step_exps}
    \end{figure}

    \subsection{Single-Step Goals} 

    Before we leverage the LLM for longer-horizon shape crafting goals, we first need to evaluate how well an LLM can out-of-the-box select the single action to achieve a single-step goal. While this is a relatively simple task, it proved somewhat difficult for the LLM due to how it is reasoning about the clay molding task. We found that the LLM prescribes semantically meaningful descriptions for the clay states and goals in order to generate action sequences. This is a very useful attribute for the long horizon shape crafting task, but can interfere with performance for the single-step goals in which the goal shape isn't anything in particular. We found the LLM was able to successfully reach the goal shape, but could sometimes get confused about rotation. The numerical results for the single-step experiments are shown in Table \ref{tab:single_step_performance}, and the qualitative results are visualized in Figure \ref{fig:single_step_exps}. For these single-step experiments, we also explored an alternative state representation that is compatible with the grid-based action space. We implemented a baseline in which we represent the clay state as a binary array with a 1 if more than half of the cell contains clay and a 0 if less than half the cell contains clay. While both this and the gridded image representations provide sufficient information for the crafting task, we argue that the image-based representation allows for shape-level reasoning more easily. This is reflected in the experimental results in which the gridded image is visually able to more consistently match the goal shape. While the LLM can get confused about rotation with both representations, this happens less frequently with the gridded image. The image representation inherently has much more information than the simple binary array.

    \begin{figure*}
    \centering
    \includegraphics[width=1.0\linewidth]{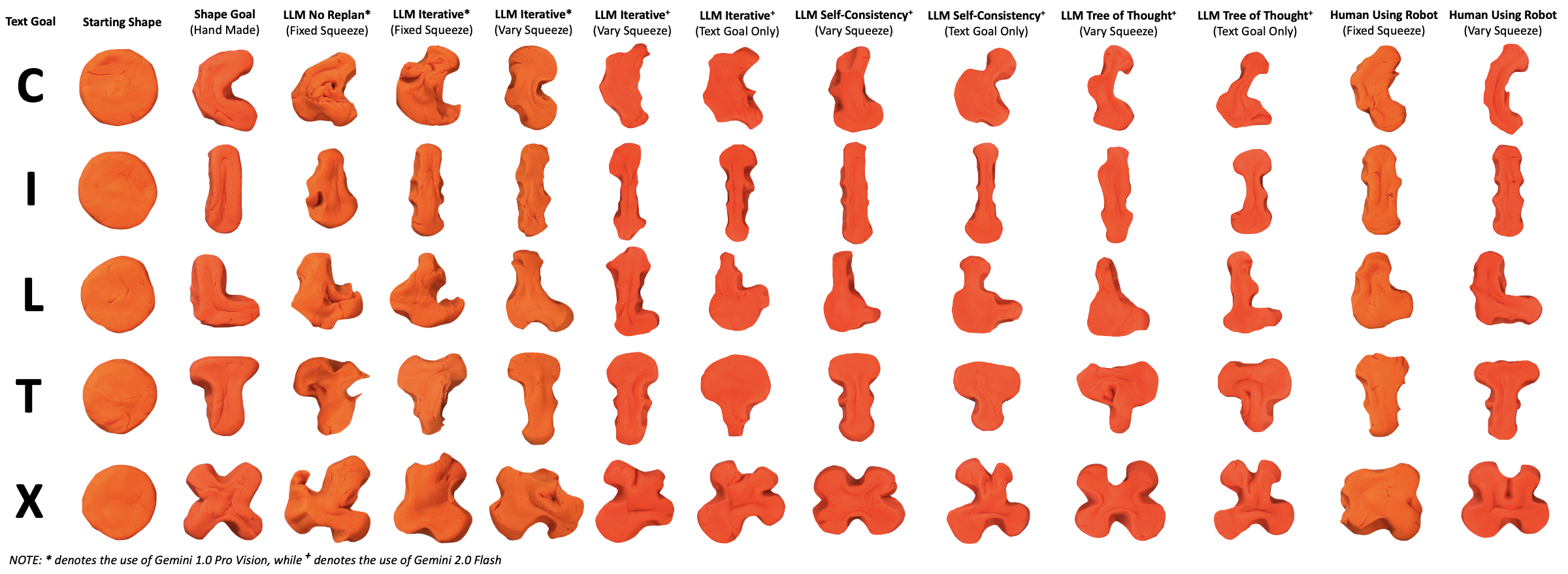}
    \vspace{-10pt}
    \caption{\textbf{Qualitative Shape Results.} We compare various LLM rollout strategies (no replanning, iterative, SC, ToT), the case of controlling the strength of the squeeze, semantic sculpting versus a ground truth goal image, and a human oracle.}
    \label{fig:shape_goals}
    \end{figure*}

    \begin{table*}[]
    \caption{\textbf{Quantitative Shape Results.} We present the mean and standard deviation across the shape goals of \textit{C, I, L, T, X}.}\label{tab:general_performance} 
    \centering
    \begin{tabular}{@{\extracolsep{\fill}}lllllllllll}
            \toprule
             \textbf{ } & \textbf{Method} & \textbf{Squeeze} & \textbf{G. Img?} & \textbf{CD [mm] $\downarrow$} & \textbf{EMD [mm] $\downarrow$} & \textbf{Curv. $\downarrow$} & \textbf{PAR $\downarrow$} & \textbf{H. Cls. $\uparrow$} & \textbf{H. Qual. $\uparrow$}  \\ 
            \midrule
            \hline
            \multirow{3}{*}{\textbf{Gemini 1.0}} & No Replan & Fixed & Yes & 5.6 $\pm$ 1.9 & 8.2 $\pm$ 2.6  & 0.089 $\pm$ 0.012 & \textbf{0.047 $\pm$ 0.002} & 0.63 $\pm$ 0.17 & 4.4 $\pm$ 1.5 \\
               & Iterative & Fixed & Yes  & \textbf{4.5 $\pm$ 1.5} & \textbf{6.7 $\pm$ 2.0} & 0.099 $\pm$ 0.021 & 0.048 $\pm$ 0.006 & 0.52 $\pm$ 0.34 & \textbf{5.9 $\pm$ 1.1} \\
               & Iterative & Varied & Yes  & 5.5 $\pm$ 2.2 & 7.8 $\pm$ 2.5  & \textbf{0.088 $\pm$ 0.008} & 0.050 $\pm$ 0.006 & \textbf{0.68 $\pm$ 0.36} & 5.0 $\pm$ 1.5 \\
               \midrule
               \multirow{6}{*}{\textbf{Gemini 2.0}} & Iterative & Varied & Yes  & 3.8 $\pm$ 1.0 & \textbf{7.7 $\pm$ 2.4}  & 0.102 $\pm$ 0.008 & 0.059 $\pm$ 0.010 & 0.81 $\pm$ 0.12 & 4.9 $\pm$ 1.5 \\
               & Iterative & Varied & No  & - & -  & 0.107 $\pm$ 0.011 & 0.055 $\pm$ 0.009 & 0.74 $\pm$ 0.17 & 4.3 $\pm$ 1.1 \\
               & SC & Varied & Yes  & \textbf{3.7 $\pm$ 0.9} & 8.0 $\pm$ 1.9 & 0.095 $\pm$ 0.018 & \textbf{0.055 $\pm$ 0.007} & 0.83 $\pm$ 0.37 & \textbf{6.4 $\pm$ 1.3} \\
               & SC & Varied & No  & - & - & 0.097 $\pm$ 0.005 & 0.059 $\pm$ 0.009 & 0.83 $\pm$ 0.37 & 5.3 $\pm$ 1.2 \\
               & ToT & Varied & Yes  & 4.2 $\pm$ 1.1 & 8.5 $\pm$ 1.7 & \textbf{0.086 $\pm$ 0.014} & 0.059 $\pm$ 0.010 & \textbf{0.85 $\pm$ 0.20} & 5.4 $\pm$ 1.2 \\
               & ToT & Varied & No  & -  & -  & 0.114 $\pm$ 0.017 & 0.063 $\pm$ 0.010 & 0.85 $\pm$ 0.34 & 6.3 $\pm$ 1.3 \\
            \midrule
            \multirow{2}{*}{\textbf{Human}} & Robot & Fixed & Yes  & 5.1 $\pm$ 2.9 & 8.2 $\pm$ 3.7  & \textbf{0.090 $\pm$ 0.011} & \textbf{0.060 $\pm$ 0.010} & 0.80 $\pm$ 0.13 & 5.5 $\pm$ 1.3 \\
            & Robot & Varied & Yes  & \textbf{2.6 $\pm$ 0.6} & \textbf{6.2 $\pm$ 1.2} & 0.110 $\pm$ 0.016 & \textbf{0.060 $\pm$ 0.010} & \textbf{1.00 $\pm$ 0.00} & \textbf{7.7 $\pm$ 1.0} \\
            \bottomrule
            
        \end{tabular}
    \end{table*}

    \begin{table}[]
    \caption{\textbf{API Cost Results.} For each method with Gemini 2.0 flash across the shape goals of \textit{C, I, L, T, X}.}\label{tab:api} 
    \centering
    \begin{tabular}{@{\extracolsep{\fill}}llllll}
            \toprule
             \textbf{Method} & \textbf{G. Img?} & \textbf{Cost [USD] $\downarrow$} & \textbf{\# Actions} \\ 
            \midrule
            \hline
            \multirow{2}{*}{Iterative} & Yes & 0.0037 $\pm$ 0.0019 &  5.16 $\pm$ 2.62 \\
            & No  & 0.0039 $\pm$ 0.0013 & 5.80 $\pm$ 1.98 \\
            \midrule
            \multirow{2}{*}{Self-Consistency} & Yes & 0.0062 $\pm$ 0.0050 &  7.00 $\pm$ 4.18 \\
            & No  & 0.0036 $\pm$ 0.0002 & 7.24 $\pm$ 1.99 \\
            \midrule
            \multirow{2}{*}{Tree of Thought}  & Yes & 0.0113 $\pm$ 0.0007 &  7.60 $\pm$ 1.39 \\
            & No  & 0.0097 $\pm$ 0.0014 & 6.48 $\pm$ 2.14 \\
            \bottomrule
            
        \end{tabular}
    \end{table}

    \subsection{Long-Horizon Shape Goals}

\begin{figure}
        \centering
        \includegraphics[width=1.0\linewidth]{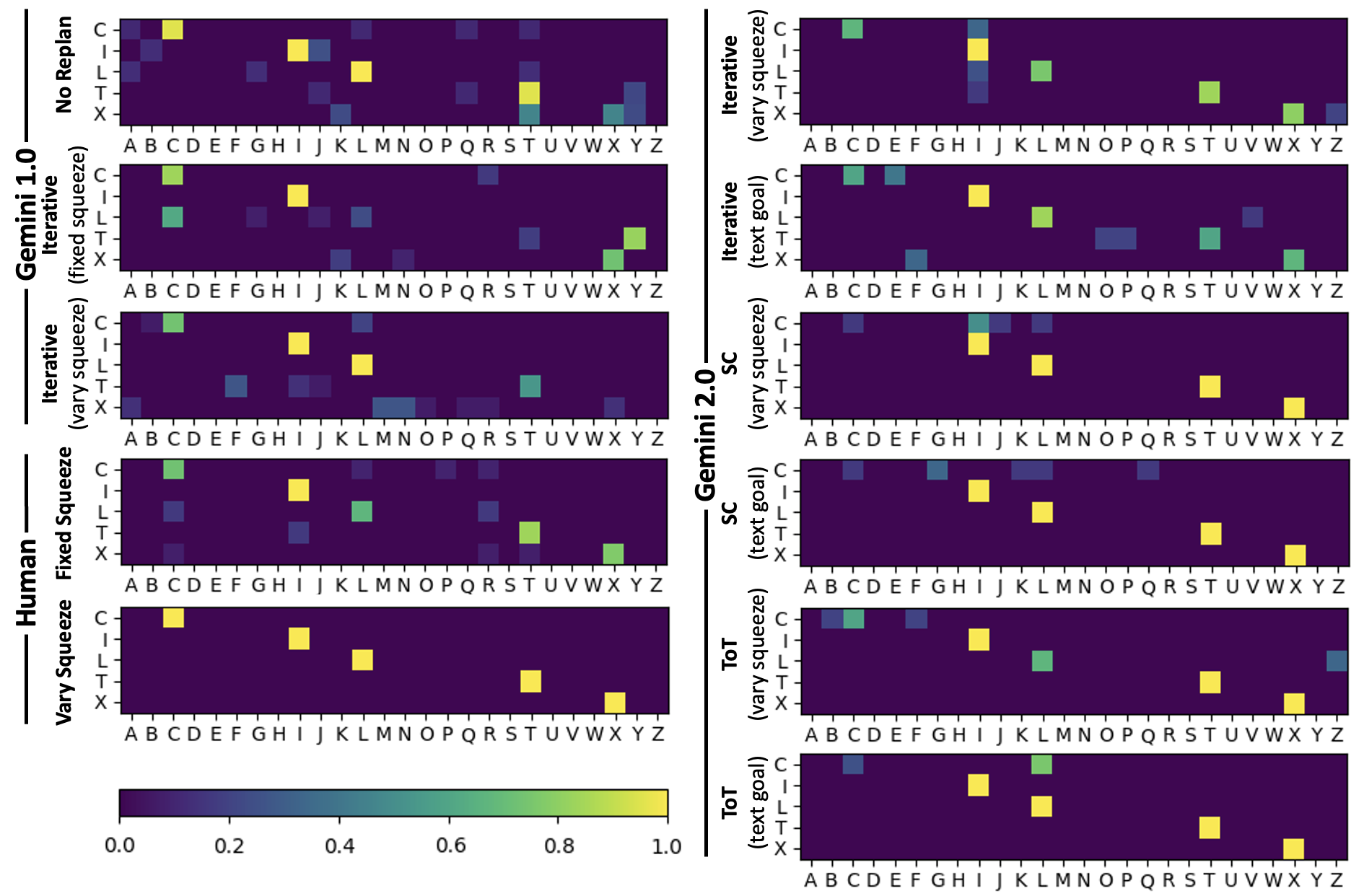}
        \vspace{-10pt}
        \caption{The confusion matrix of the human evaluations for all methods. A score of 1.0 indicates all human respondents classified the clay image as that letter.}
        \label{fig:confusion}
    \end{figure}

    To further investigate the capabilities of LLMs for crafting, we evaluated the full system performance on a set of long-horizon letter shape goals consisting of \textit{C, I, L, T, X}. We selected these letters because they each require different skills and long-horizon strategies to achieve, and are all able to be created with only deformation actions, whereas letters such as \textit{P} are not because our current action formulation cannot create holes. Through our experiments, we compare different rollout strategies (no replanning, iterative, self-consistency, and tree of thought), LLM and human action control with a fixed or varied gripper distance, and methods with and without a goal image provided. The fixed squeeze condition is one in which the robot always squeezes the end-effector to a fixed distance of 1.5cm between the fingertips. The varied squeeze condition allows the agent to select 'min', 'medium', or 'max', corresponding to 2.5cm, 1.5cm, and 0.8cm respectively. The qualitative results are visualized in Figure \ref{fig:shape_goals}, the quantitative results are shown in Table \ref{tab:general_performance}, the LLM API costs and number of actions are shown in Table \ref{tab:api}, and a visualization of the human classification results across all methods are shown in Figure \ref{fig:confusion}.

    Through our experiments, we find that the LLM-Craft framework is able to successfully create visually identifiable shapes. In particular, we find that allowing the LLM to control both the gripper position and squeeze strength improves the quality of the final shapes created, particularly in terms of the human evaluations. Self-Consistency (SC) allowed for a selection of more consistent grasp strengths, which can be exemplified in the \textit{I} shape results. Tree of thought (ToT) and SC rollout methods both outperform the iterative strategy in terms of shape smoothness, and human evaluation results. SC and ToT rollout strategies are able to more reliably generate nuanced sculpting trajectories, such as squeezing maximum strength on the first grasp and medium strength on the second perpendicular grasp to better define the final \textit{X} shape without pushing the arms together. The improvement to shape quality by these sophisticated rollout methods can be best visualized for the \textit{X} shape in Figure \ref{fig:shape_goals}, and in improvements in human classification and quality ratings as compared to the iterative case. SC and ToT consistently are rated higher in terms of classification and quality (Welch's t-test, $p_{SC}=0.06$, $t_{SC}=2.4$, $p_{ToT}=0.01$, $t_{ToT} = 3.6$). Across the shapes, the human oracle with the ability to vary the squeeze of the gripper outperforms the best performing LLM-Craft variants of SC and ToT in quality rating by 1.3 and 2.3 respectfully (Welch's t-test, $p_{SC}=0.006$, $t_{SC}=2.9$, $p_{ToT}=0.003$, $t_{ToT} = 3.2$).

    We find that CD/EMD are unreliable quantitative metrics for determining small differences in shape quality between our method variants. The CD/EMD  metrics do not align well with the human evaluation results. This is because CD and EMD substantially penalize slight misalignment in terms of rotation, or shape structure with the specific goal point cloud, whereas the human raters prioritize shape recognizability as the goal letter and general visible smoothness. For the purpose of creating a semantically meaningful and recognizable shape, not all points on the surface are created equal. Certain deviations from the goal have little impact on human classification, whereas others have a significant impact. Some attributes of the shape are more important than others in terms of final classification and quality results. We hope to further explore and develop additional quantitative metrics that better align with human evaluations in future work.

    The shape curvature and perimeter-to-area ratio are not quantitative metrics that alone can distinguish how well a created shape matches the goal. However, they do allow us to analyze nuanced differences between methods that perform similarly in terms of CD/EMD as well as human classification/quality ratings. In particular, while the SC method with and without a goal image have comparable human classification and quality rating results, the SC method with a goal image are "smoother" in terms of lower scores in both the curvature and perimeter-to-area ratio across all shapes. Similarly, for ToT, the method with a goal image has lower curvature and perimeter-to-area ratio scores across all shapes. Both SC and ToT rollout methods consistently created letter shapes that were recognizable by human evaluators, with classification accuracy outperforming all other non-human methods (visualized in Figure \ref{fig:confusion}). The primary shape these methods struggled to create, and human evaluators then struggled to classify was the \textit{C} shape. This disparity between the \textit{C} shape and the other letters is reflected in the larger standard deviations in the human evaluations. For a letter-by-letter breakdown of human evaluation results, please see the supplementary materials on the project website. Overall, the \textit{C} shape was the most difficult shape for our system to create, largely due to the discrete action space. The \textit{C} shape requires a smooth, tight curve which is very difficult to create with a discrete action grid. The challenge of this is reflected in the quality of \textit{C} shapes the human oracle was able to create both with fixed/varied squeezes.

\begin{table}[]
\caption{\textbf{Quantitative Grid Ablation Results.} We present the mean and standard deviation for the \textit{X} shape with varying grids.}\label{tab:grid} 
\centering
\begin{tabular}{@{\extracolsep{\fill}}llllll}
        \hline
        \textbf{} & \textbf{} & \textbf{H. Qual. $\uparrow$} & \textbf{Curv. $\downarrow$} & \textbf{PAR $\downarrow$} \\ 
        \hline
        \hline
        \multirow{2}{*}{\textbf{6x6}} & Iterative & 4.6 $\pm$ 2.6 & 0.059 $\pm$ 0.005 & 0.029 $\pm$ 0.003  \\
          & Human & 3.6 $\pm$ 2.5 & 0.074 $\pm$ 0.016 & 0.028 $\pm$ 0.001  \\
        \hline
        \multirow{2}{*}{\textbf{8x8}} & Iterative & 3.8 $\pm$ 3.1 & 0.064 $\pm$ 0.006 & 0.030 $\pm$ 0.001 \\
          & Human & 4.2 $\pm$ 2.7 & 0.070 $\pm$ 0.003 & 0.030 $\pm$ 0.003 \\
        \hline
        \multirow{2}{*}{\textbf{10x10}} & Iterative & 4.8 $\pm$ 2.4 & 0.068 $\pm$ 0.011 & 0.029 $\pm$ 0.003 \\
          & Human & 4.0 $\pm$ 2.9 & 0.073 $\pm$ 0.003 & 0.030 $\pm$ 0.003 \\
        \hline
        \multirow{2}{*}{\textbf{12x12}} & Iterative & 3.2 $\pm$ 3.3 & 0.077 $\pm$ 0.005 & 0.033 $\pm$ 0.003 \\
          & Human & 3.0 $\pm$ 2.3 & 0.074 $\pm$ 0.009 & 0.034 $\pm$ 0.002 \\
        \hline
        \multirow{2}{*}{\textbf{16x16}} & Iterative & 4.2 $\pm$ 2.9 & 0.085 $\pm$ 0.006 & 0.034 $\pm$ 0.001 \\
          & Human & 5.6 $\pm$ 2.4 & 0.081 $\pm$ 0.016 & 0.033 $\pm$ 0.004  \\
        \hline
    \end{tabular}
\end{table}

    \subsection{Grid Ablation}
    \label{sec:grid}

    \begin{figure}
    \centering
    \includegraphics[width=1.0\linewidth]{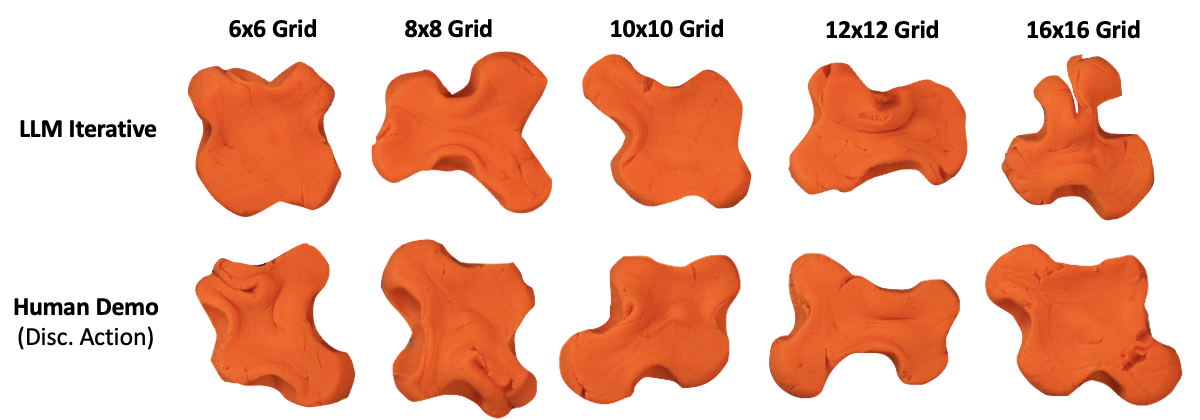}
    \vspace{-10pt}
    \caption{\textbf{Qualitative Grid Ablation.} We compare the performance of LLM-Craft compared to a human baseline for variable grid sizes. }
    \label{fig:grid}
    \end{figure}

    To evaluate if the proposed framework can scale successfully with different grid sizes, we deployed the LLM-Craft iterative pipeline on hardware for the \textit{X} letter task with grids of 6x6, 8x8, 10x10, 12x12 and 16x16. We conduct 5 experimental runs for each grid and present the visualization of the shape results in Figure \ref{fig:grid} and the quantitative results in Table \ref{tab:grid}. Additionally, we extended the human baseline in which a human selects actions with the same discrete action space to better visualize the shape capabilities of the constrained state/action space. For this ablation experiment, we kept the 3D printed fingertips the same across experiments, meaning that varying the grid size allowed for finer control of the end-effector position, but the same quantity of clay was being moved with each squeeze. Through these experiments, we find that LLM-Craft can scale to grid sizes of 6x6, 8x8, and 10x10, with comparable quality, curvature and perimeter-to-area ratio mean values. However, when the grid gets particularly large (around 12x12 or 16x16), it becomes much more difficult to select quality actions, and LLM-Craft creates worse quality shapes.

    \subsection{Prompt Ablation} 
    \label{sec:ablation}

\begin{figure}
    \centering
       \includegraphics[width=1.0\linewidth]{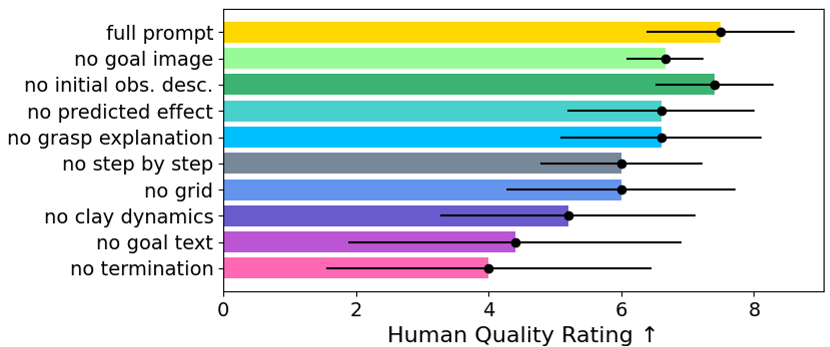}
       \vspace{-10pt}
    \caption{\textbf{Prompt Ablation.} The performance of the long-horizon, iterative system on the \textit{X} shape task as we remove different components of the prompt. For each prompt variation, we conducted 5 hardware experimental runs. We report the mean human quality evaluation, with the black bar indicating the standard deviation.}
    \label{fig:ablation}
    \end{figure}

    To evaluate the impact of different components of the prompts on the performance, we conducted an ablation study removing different parts of the prompt. We conducted a second human evaluation survey for this ablation study in which we provided the evaluators with an image of the \textit{X} shape created by each prompt variant. In this case, the evaluators knew that the shape was supposed to be an \textit{X}, and were asked to rate the quality of the \textit{X} shape on a scale of 1-10. In particular, we chose to remove the following (the order reflects the order shown in Figure \ref{fig:ablation}): \textbf{(1)} goal image, \textbf{(2)} asking the LLM to describe the similarities and differences between the state and the goal, \textbf{(3)} asking the LLM to predict the effect of the grasp on the clay after selecting an action, \textbf{(4)} the explanation of grasp types, specifically horizontal, vertical, or diagonal, \textbf{(5)} the instructions to choose actions step-by-step, \textbf{(6)} the explanations of the gridded state and action space, \textbf{(7)} explaining the clay behavior when a cell is squeezed, \textbf{(8)} the goal text description (i.e. 'X'), and \textbf{(9)} the termination prompt. The breakdown of the human evaluations for the \textit{X} shape across the ablations in shown in Figure \ref{fig:ablation}. 

    Based on our ablation experiments, the removal of the termination module negatively impacts performance the most. This is an expected result, because without the ability to recognize when the goal shape has been reached, the agent may continue to squeeze the clay and worsen the final shape created. The explanation of clay dynamics and the inclusion of the goal in text form have a large impact on the final shape quality. The explanation of clay dynamics is particularly important, as it explains how to conceptualize the changes in the cell-based state after a squeeze action. There was a medium impact on the final shape quality when the explanation of the grid and the step-by-step requirements were removed from the prompt. The reasoning about the predicted effect, grasp explanation, and the goal image had a small impact to the final shape quality. Finally, the requirement to reason about the state and goal had no effect on the final shape quality compared to the full prompt.

    \subsection{Semantic Goals}

    In general, much of past literature for elaso-plastic object robotic crafting has focused on explicit shape goals \cite{bartsch2023sculptbot, bartsch2024sculptdiff, shi2022robocraft, bauer2024doughnet, shi2023robocook}. However, the core goal of creating a sculpture is to represent a shape or an idea, not necessarily exactly replicate an existing sculpture or object. These more semantic goals are incredibly difficult to capture, quantify and evaluate. However, given that we are using language in the LLM-Craft framework, we have a system that is more aligned with understanding and creating these semantic-based goals. For example, the termination prompt is evaluating more abstract shape similarity as opposed to low-level point similarity. To evaluate how well our framework can handle these semantic goals, we conducted a variation of each rollout method in which we removed the goal image from the prompt. The results of these variants are shown in Figure \ref{fig:shape_goals} and Table \ref{tab:general_performance}. We found that the LLM-Craft framework was indeed still able to create the simple letter shapes with a text-based goal alone. Without a goal image for guidance, the iterative framework would often stop earlier, whenever the termination agent would initially identify the letter in the clay. This resulted in an average of fewer actions, and final shapes that are recognized less often as compared to the iterative rollout with goal images (0.74 versus 0.81 mean classification accuracy, respectively). However, this discrepancy is resolved with the more sophisticated rollout strategies of SC (0.82 mean classification accuracy for both) and ToT (0.85 mean classification accuracy for both). Interestingly, the methods with text-based goals also have a wider variety of instances of the letters as compared to those with an explicit goal image. For example, the \textit{I} shape created by Tree of Thought without a goal image includes the small horizontal top/bottom lines on the \textit{I} shape that appear in some fonts. ToT without a goal image opts to squeeze the two most inward cells, leaving the small horizontal top/bottom lines on the \textit{I} shape that appear in some fonts. While these are relatively simple semantically meaningful shapes, we hope to extend this exploration of semantic goals for 3D sculpture in future work.

\section{Conclusion}
\label{sec:conclusion}

    In this work, we present LLM-Craft, to the best of our knowledge the first LLM-based robotic shape crafting system. Through real-world experiments, we demonstrate the powerful planning capabilities of LLMs out-of-the-box for both single-step and longer horizon shape goals. We find that LLMs contain relevant world knowledge for the clay crafting task without requiring any task-specific data, such as when creating an \textit{I} one must squeeze along a single axis, versus when creating an \textit{X} one must squeeze perpendicular to the previous grasp. For broader applications, our results demonstrate the usefulness of incorporating LLMs into robotic domains that typically require datasets that are task-specific, hardware-specific, and difficult to acquire. Additionally, we find that implementing more sophisticated rollout strategies such as Self-Consistency and Tree of Thought improves performance further and, in particular, reduces the frequency in which the system enters an irrecoverable clay state. We believe that the world knowledge and semantic reasoning capabilities of LLMs could be very useful in future deformable object manipulation work that combines the more classical and physics-based approaches with LLMs to reduce real-world dataset needs, and incorporate LLMs for higher-level reasoning and semantic similarity analysis. 

    Although our results demonstrate that LLM-Craft is effective in creating simple letter shapes, there is a clear limitation in shape complexity due to the grid-based action space. In particular, our framework struggles with the \textit{C} shape because it is difficult to create a smooth, tight curve with a discrete, grid-based action space. We would expect similar limitations if LLM-Craft is extended to more complex shapes without developing a continuous action space. This framework could be extended to a continuous top-down action space by employing iterative visual prompting strategies similar to those in \cite{nasiriany2024pivot}. Furthermore, this framework could be extended to fully 3D shapes by incorporating multiple camera views, or depth images. We hope to explore these approaches in future work.


\bibliographystyle{IEEEtran}
\bibliography{ref}

\begin{thebibliography}{10}
\providecommand{\url}[1]{#1}
\csname url@samestyle\endcsname
\providecommand{\newblock}{\relax}
\providecommand{\bibinfo}[2]{#2}
\providecommand{\BIBentrySTDinterwordspacing}{\spaceskip=0pt\relax}
\providecommand{\BIBentryALTinterwordstretchfactor}{4}
\providecommand{\BIBentryALTinterwordspacing}{\spaceskip=\fontdimen2\font plus
\BIBentryALTinterwordstretchfactor\fontdimen3\font minus \fontdimen4\font\relax}
\providecommand{\BIBforeignlanguage}[2]{{%
\expandafter\ifx\csname l@#1\endcsname\relax
\typeout{** WARNING: IEEEtran.bst: No hyphenation pattern has been}%
\typeout{** loaded for the language `#1'. Using the pattern for}%
\typeout{** the default language instead.}%
\else
\language=\csname l@#1\endcsname
\fi
#2}}
\providecommand{\BIBdecl}{\relax}
\BIBdecl

\bibitem{shi2022robocraft}
H.~Shi, H.~Xu, Z.~Huang, Y.~Li, and J.~Wu, ``Robocraft: Learning to see, simulate, and shape elasto-plastic objects in 3d with graph networks,'' \emph{IJRR}, vol.~43, no.~4, pp. 533--549, 2024.

\bibitem{bartsch2023sculptbot}
A.~Bartsch, C.~Avra, and A.~B. Farimani, ``Sculptbot: Pre-trained models for 3d deformable object manipulation,'' in \emph{ICRA}.\hskip 1em plus 0.5em minus 0.4em\relax IEEE, 2024, pp. 12\,548--12\,555.

\bibitem{shi2023robocook}
H.~Shi, H.~Xu, S.~Clarke, Y.~Li, and J.~Wu, ``Robocook: Long-horizon elasto-plastic object manipulation with diverse tools,'' \emph{CoRL}, 2023.

\bibitem{bartsch2024sculptdiff}
A.~Bartsch, A.~Car, C.~Avra, and A.~B. Farimani, ``Sculptdiff: Learning robotic clay sculpting from humans with goal conditioned diffusion policy,'' in \emph{IROS}.\hskip 1em plus 0.5em minus 0.4em\relax IEEE, 2024, pp. 7307--7314.

\bibitem{bauer2024doughnet}
D.~Bauer, Z.~Xu, and S.~Song, ``Doughnet: A visual predictive model for topological manipulation of deformable objects,'' in \emph{ECCV}.\hskip 1em plus 0.5em minus 0.4em\relax Springer, 2024, pp. 92--108.

\bibitem{mirchandani2023large}
S.~Mirchandani, F.~Xia, P.~Florence, B.~Ichter, D.~Driess, M.~G. Arenas, K.~Rao, D.~Sadigh, and A.~Zeng, ``Large language models as general pattern machines,'' \emph{preprint arXiv:2307.04721}, 2023.

\bibitem{wu2023smartplay}
Y.~Wu, X.~Tang, T.~M. Mitchell, and Y.~Li, ``Smartplay: A benchmark for llms as intelligent agents,'' \emph{preprint arXiv:2310.01557}, 2023.

\bibitem{todd2023level}
G.~Todd, S.~Earle, M.~U. Nasir, M.~C. Green, and J.~Togelius, ``Level generation through large language models,'' in \emph{FDG}, 2023, pp. 1--8.

\bibitem{yu2023language}
W.~Yu, N.~Gileadi, C.~Fu, S.~Kirmani, K.-H. Lee, M.~G. Arenas, H.-T.~L. Chiang, T.~Erez, L.~Hasenclever, J.~Humplik \emph{et~al.}, ``Language to rewards for robotic skill synthesis,'' \emph{preprint arXiv:2306.08647}, 2023.

\bibitem{jadhav2024large}
Y.~Jadhav and A.~B. Farimani, ``Large language model agent as a mechanical designer,'' \emph{preprint arXiv:2404.17525}, 2024.

\bibitem{wang2024can}
S.~Wang, Z.~Wei, Y.~Choi, and X.~Ren, ``Can llms reason with rules? logic scaffolding for stress-testing and improving llms,'' \emph{preprint arXiv:2402.11442}, 2024.

\bibitem{singh2023progprompt}
I.~Singh, V.~Blukis, A.~Mousavian, A.~Goyal, D.~Xu, J.~Tremblay, D.~Fox, J.~Thomason, and A.~Garg, ``Progprompt: program generation for situated robot task planning using large language models,'' \emph{Autonomous Robots}, vol.~47, no.~8, pp. 999--1012, 2023.

\bibitem{sharan2024plan}
S.~Sharan, R.~Zhao, Z.~Wang, S.~P. Chinchali \emph{et~al.}, ``Plan diffuser: Grounding llm planners with diffusion models for robotic manipulation,'' in \emph{Bridging the Gap between Cognitive Science and Robot Learning in the Real World: Progresses and New Directions}, 2024.

\bibitem{wu2024mldt}
Y.~Wu, J.~Zhang, N.~Hu, L.~Tang, G.~Qi, J.~Shao, J.~Ren, and W.~Song, ``Mldt: Multi-level decomposition for complex long-horizon robotic task planning with open-source large language model,'' \emph{preprint arXiv:2403.18760}, 2024.

\bibitem{bhat2024grounding}
V.~Bhat, A.~U. Kaypak, P.~Krishnamurthy, R.~Karri, and F.~Khorrami, ``Grounding llms for robot task planning using closed-loop state feedback,'' \emph{preprint arXiv:2402.08546}, 2024.

\bibitem{gupta2024action}
S.~Gupta, K.~Yao, L.~Niederhauser, and A.~Billard, ``Action contextualization: Adaptive task planning and action tuning using large language models,'' \emph{preprint arXiv:2404.13191}, 2024.

\bibitem{wei2022chain}
J.~Wei, X.~Wang, D.~Schuurmans, M.~Bosma, F.~Xia, E.~Chi, Q.~V. Le, D.~Zhou \emph{et~al.}, ``Chain-of-thought prompting elicits reasoning in large language models,'' \emph{NeurIPS}, vol.~35, pp. 24\,824--24\,837, 2022.

\bibitem{wang2022self}
X.~Wang, J.~Wei, D.~Schuurmans, Q.~Le, E.~Chi, S.~Narang, A.~Chowdhery, and D.~Zhou, ``Self-consistency improves chain of thought reasoning in language models,'' \emph{preprint arXiv:2203.11171}, 2022.

\bibitem{yao2023tree}
S.~Yao, D.~Yu, J.~Zhao, I.~Shafran, T.~Griffiths, Y.~Cao, and K.~Narasimhan, ``Tree of thoughts: Deliberate problem solving with large language models,'' \emph{NeurIPS}, vol.~36, pp. 11\,809--11\,822, 2023.

\bibitem{heiden2021disect}
E.~Heiden, M.~Macklin, Y.~Narang, D.~Fox, A.~Garg, and F.~Ramos, ``Disect: A differentiable simulation engine for autonomous robotic cutting,'' \emph{preprint arXiv:2105.12244}, 2021.

\bibitem{hu2018moving}
Y.~Hu, Y.~Fang, Z.~Ge, Z.~Qu, Y.~Zhu, A.~Pradhana, and C.~Jiang, ``A moving least squares material point method with displacement discontinuity and two-way rigid body coupling,'' \emph{ACM TOG}, vol.~37, no.~4, pp. 1--14, 2018.

\bibitem{huang2021plasticinelab}
Z.~Huang, Y.~Hu, T.~Du, S.~Zhou, H.~Su, J.~B. Tenenbaum, and C.~Gan, ``Plasticinelab: A soft-body manipulation benchmark with differentiable physics,'' \emph{preprint arXiv:2104.03311}, 2021.

\bibitem{gu2023maniskill2}
J.~Gu, F.~Xiang, X.~Li, Z.~Ling, X.~Liu, T.~Mu, Y.~Tang, S.~Tao, X.~Wei, Y.~Yao \emph{et~al.}, ``Maniskill2: A unified benchmark for generalizable manipulation skills,'' \emph{preprint arXiv:2302.04659}, 2023.

\bibitem{qi2022learning}
C.~Qi, X.~Lin, and D.~Held, ``Learning closed-loop dough manipulation using a differentiable reset module,'' \emph{RA-L}, vol.~7, no.~4, pp. 9857--9864, 2022.

\bibitem{yamada2024d}
J.~Yamada, S.~Zhong, J.~Collins, and I.~Posner, ``D-cubed: Latent diffusion trajectory optimisation for dexterous deformable manipulation,'' \emph{preprint arXiv:2403.12861}, 2024.

\bibitem{li2024deformnet}
C.~Li, Z.~Ai, T.~Wu, X.~Li, W.~Ding, and H.~Xu, ``Deformnet: Latent space modeling and dynamics prediction for deformable object manipulation,'' \emph{preprint arXiv:2402.07648}, 2024.

\bibitem{lin2022planning}
X.~Lin, C.~Qi, Y.~Zhang, Z.~Huang, K.~Fragkiadaki, Y.~Li, C.~Gan, and D.~Held, ``Planning with spatial-temporal abstraction from point clouds for deformable object manipulation,'' \emph{preprint arXiv:2210.15751}, 2022.

\bibitem{li2023dexdeform}
S.~Li, Z.~Huang, T.~Chen, T.~Du, H.~Su, J.~B. Tenenbaum, and C.~Gan, ``Dexdeform: Dexterous deformable object manipulation with human demonstrations and differentiable physics,'' \emph{preprint arXiv:2304.03223}, 2023.

\bibitem{wang2023describe}
Z.~Wang, S.~Cai, G.~Chen, A.~Liu, X.~Ma, and Y.~Liang, ``Describe, explain, plan and select: Interactive planning with large language models enables open-world multi-task agents,'' \emph{preprint arXiv:2302.01560}, 2023.

\bibitem{rana2023sayplan}
K.~Rana, J.~Haviland, S.~Garg, J.~Abou-Chakra, I.~Reid, and N.~Suenderhauf, ``Sayplan: Grounding large language models using 3d scene graphs for scalable task planning,'' \emph{preprint arXiv:2307.06135}, 2023.

\bibitem{dorbala2023can}
V.~S. Dorbala, J.~F. Mullen~Jr, and D.~Manocha, ``Can an embodied agent find your “cat-shaped mug”? llm-based zero-shot object navigation,'' \emph{RA-L}, 2023.

\bibitem{ding2023task}
Y.~Ding, X.~Zhang, C.~Paxton, and S.~Zhang, ``Task and motion planning with large language models for object rearrangement,'' in \emph{IROS}.\hskip 1em plus 0.5em minus 0.4em\relax IEEE, 2023, pp. 2086--2092.

\bibitem{nasiriany2024pivot}
S.~Nasiriany, F.~Xia, W.~Yu, T.~Xiao, J.~Liang, I.~Dasgupta, A.~Xie, D.~Driess, A.~Wahid, Z.~Xu \emph{et~al.}, ``Pivot: Iterative visual prompting elicits actionable knowledge for vlms,'' \emph{preprint arXiv:2402.07872}, 2024.

\bibitem{cheng2024empowering}
G.~Cheng, C.~Zhang, W.~Cai, L.~Zhao, C.~Sun, and J.~Bian, ``Empowering large language models on robotic manipulation with affordance prompting,'' \emph{preprint arXiv:2404.11027}, 2024.

\bibitem{wang2024grounding}
Y.~Wang, T.-H. Wang, J.~Mao, M.~Hagenow, and J.~Shah, ``Grounding language plans in demonstrations through counterfactual perturbations,'' \emph{preprint arXiv:2403.17124}, 2024.

\bibitem{driess2023palm}
D.~Driess, F.~Xia, M.~S. Sajjadi, C.~Lynch, A.~Chowdhery, B.~Ichter, A.~Wahid, J.~Tompson, Q.~Vuong, T.~Yu \emph{et~al.}, ``Palm-e: An embodied multimodal language model,'' \emph{preprint arXiv:2303.03378}, 2023.

\bibitem{huang2023voxposer}
W.~Huang, C.~Wang, R.~Zhang, Y.~Li, J.~Wu, and L.~Fei-Fei, ``Voxposer: Composable 3d value maps for robotic manipulation with language models,'' \emph{preprint arXiv:2307.05973}, 2023.

\bibitem{liang2023code}
J.~Liang, W.~Huang, F.~Xia, P.~Xu, K.~Hausman, B.~Ichter, P.~Florence, and A.~Zeng, ``Code as policies: Language model programs for embodied control,'' in \emph{ICRA}.\hskip 1em plus 0.5em minus 0.4em\relax IEEE, 2023, pp. 9493--9500.

\bibitem{mikami2024natural}
Y.~Mikami, A.~Melnik, J.~Miura, and V.~Hautam{\"a}ki, ``Natural language as polices: Reasoning for coordinate-level embodied control with llms,'' \emph{preprint arXiv:2403.13801}, 2024.

\bibitem{shah2023lm}
D.~Shah, B.~Osi{\'n}ski, S.~Levine \emph{et~al.}, ``Lm-nav: Robotic navigation with large pre-trained models of language, vision, and action,'' in \emph{CoRL}.\hskip 1em plus 0.5em minus 0.4em\relax PMLR, 2023, pp. 492--504.

\bibitem{yang2023llm}
J.~Yang, X.~Chen, S.~Qian, N.~Madaan, M.~Iyengar, D.~F. Fouhey, and J.~Chai, ``Llm-grounder: Open-vocabulary 3d visual grounding with large language model as an agent,'' \emph{preprint arXiv:2309.12311}, 2023.

\bibitem{wu2023tidybot}
J.~Wu, R.~Antonova, A.~Kan, M.~Lepert, A.~Zeng, S.~Song, J.~Bohg, S.~Rusinkiewicz, and T.~Funkhouser, ``Tidybot: Personalized robot assistance with large language models,'' \emph{Autonomous Robots}, vol.~47, no.~8, pp. 1087--1102, 2023.

\bibitem{xu2023creative}
M.~Xu, P.~Huang, W.~Yu, S.~Liu, X.~Zhang, Y.~Niu, T.~Zhang, F.~Xia, J.~Tan, and D.~Zhao, ``Creative robot tool use with large language models,'' \emph{preprint arXiv:2310.13065}, 2023.

\bibitem{kwon2023language}
T.~Kwon, N.~Di~Palo, and E.~Johns, ``Language models as zero-shot trajectory generators,'' in \emph{Workshop LangRob}, 2023.

\bibitem{team2023gemini}
G.~Team, R.~Anil, S.~Borgeaud, Y.~Wu, J.-B. Alayrac, J.~Yu, R.~Soricut, J.~Schalkwyk, A.~M. Dai, A.~Hauth \emph{et~al.}, ``Gemini: a family of highly capable multimodal models,'' \emph{preprint arXiv:2312.11805}, 2023.

\bibitem{Pichai_Hassabis_Kavukcuoglu_2024}
\BIBentryALTinterwordspacing
S.~Pichai, D.~Hassabis, and K.~Kavukcuoglu, ``Introducing gemini 2.0: Our new ai model for the agentic era,'' Dec 2024. [Online]. Available: \url{https://blog.google/technology/google-deepmind/google-gemini-ai-update-december-2024/}
\BIBentrySTDinterwordspacing

\bibitem{achiam2023gpt}
J.~Achiam, S.~Adler, S.~Agarwal, L.~Ahmad, I.~Akkaya, F.~L. Aleman, D.~Almeida, J.~Altenschmidt, S.~Altman, S.~Anadkat \emph{et~al.}, ``Gpt-4 technical report,'' \emph{preprint arXiv:2303.08774}, 2023.

\end{thebibliography}

\end{document}